\pdfoutput=1
\documentclass[11pt]{article}

\usepackage[]{EMNLP2023}

\usepackage{times}
\usepackage{latexsym}
\usepackage[T1]{fontenc}
\usepackage[utf8]{inputenc}

\usepackage[verbose=silent]{microtype}

\usepackage{inconsolata}

\usepackage{tabularx}
\usepackage{dingbat}
\usepackage{linguex}
\usepackage{float}
\usepackage{multirow}
\usepackage{url}
\usepackage{booktabs}
\usepackage{subcaption}
\usepackage[normalem]{ulem}
\usepackage[final]{proofing} 

\newcommand{\ST}{\textit{Self-Contained Neg Test}}

\title{The Self-Contained Negation Test Set}

\author{David Kletz$^{1,2}$  \and Pascal Amsili$^2$ \and Marie Candito$^1$\\
         (1) Université Paris Cité \& LLF (CNRS/UPC)\\ (2) Université Sorbonne Nouvelle \& Lattice (CNRS/ENS-PSL/USN)\\ {\tt\small david.kletz@sorbonne-nouvelle.fr}, {\tt\small marie.candito@u-paris.fr}, {\tt\small pascal.amsili@ens.fr}}

\date{}

\begin{document}
\maketitle

\begin{abstract}
Several methodologies have recently been proposed to evaluate the ability of Pretrained Language Models (PLMs) to interpret negation.
In this article, we build on \citet{gubelmann-handschuh-2022-context}, which studies the modification of PLMs' predictions as a function of the polarity of inputs, in English.
%marie: reformulé infra: This test is self-contained: it relies on a pair of sentences, the polarity of the first of which constrains the predictions that can be made by a model to replace a mask in the second.
Crucially, this test uses ``self-contained'' inputs ending with a masked position: depending on the polarity of a verb in the input, a particular token is either semantically ruled out or allowed at the masked position.
%marie: reformulé infra:Nevertheless, the examples used are not aligned, and some of them are ineffective for model evaluation.   
By replicating \citet{gubelmann-handschuh-2022-context} experiments, we have uncovered flaws that weaken the conclusions that can be drawn from this test.
%To remedy these shortcomings, we're introducing the \ST, a test that extends the \citet{gubelmann-handschuh-2022-context} test, but is better controlled and systematized so as to evaluate precisely only the impact of negation.
%les deux versions étaient restées dans le texte. J'ai conservé la seconde.
We thus propose an improved version, the \ST, which is more controlled, more systematic, and entirely based on examples forming minimal pairs varying only in the presence or absence of verbal negation in English.

%a test that is an extension of the \cite{gubelmann-handschuh-2022-context} test, but better controlled and systematized so as to precisely evaluate only the impact of negation.
%For this test, we only use examples forming minimal pairs; moreover, we make sure to control the other linguistic phenomena involved, in particular those linked to coreference resolution.
% marie: "remarkable understanding" : je suis pas d'accord cf. encore bcp d'exemples où la négation amène à la répétition interdite
When applying our test to the \texttt{roberta} and \texttt{bert} base and large models, we show that only \texttt{roberta-large} shows trends that match the expectations, while \texttt{bert-base} is mostly insensitive to negation. 
%The application of this test reveals that, while the \texttt{roberta-large} model shows a remarkable understanding of negation, the \texttt{roberta-base} and \texttt{bert-large} models show a notable lack of consideration for negation in their predictions, while \texttt{bert-base} shows total insensitivity to its presence.
For all the tested models though, in a significant number of test instances the top-1 prediction remains the token that is semantically forbidden by the context, which shows how much room for improvement remains for a proper treatment of the negation phenomenon.
\end{abstract}

\section{Introduction}

The treatment of negation by PLMs has recently been the subject of various works whose conclusions are fairly contradictory.

On the one hand, \citet{kassner_negated_2020} and \citet{ettinger_what_2020} compare the predictions of Transformer-based language models \citep{attention_vaswani} in minimal pairs varying in polarity~\Next.

\ex. \label{ex:Robin}
\a. A robin is a [MASK].
\b. A robin is not a [MASK].

Noting that changes of polarity in the model's inputs result in little or no change for both top-1 predictions and the entire vocabulary distribution, these authors conclude that the models are insensitive to negation.

However, it has been established that the presence of negation can be detected in contextual representations. \citet{celikkanat_controlling_2020} thus find ``traces'' of negation on the negated verb, its subject, its object. Moreover, the extent to which negation is diffused in contextual representations follows syntactic constraints: \citet{kletz2023probing} show that the presence of negation in contextual representations is stronger for tokens within the scope of negation, this effect being visible even when controlling for the distance between the token and negation.

%\TBD{However}
As pointed out by \citet{gubelmann-handschuh-2022-context}, this apparent contradiction can be explained by the fact that \citet{kassner_negated_2020} study the factual knowledge of models, and therefore use contexts involving world knowledge, such as \Last. The inability of models \textit{not} to predict
%to predict anything other than 
\textit{bird} %even 
in the negated case could be explained by stored factual knowledge taking precedence over the ability to capture that negation reverses the truth value of a proposition. Especially as there is an asymmetry in the number of acceptable words to replace the mask: only a few are possible for the positive version, but a huge number are for the negative one, which can't be a favorable situation when only the top-1 prediction is studied.

% PLutot 
%\DK{\citet{gubelmann-handschuh-2022-context} took up the idea of an assessment based on sentence pairs \citep{warstadt-etal-2020-blimp-benchmark}. They have thus proposed a test where the inputs supplied to the models are self-contained (in our terminology): a context sentence is followed by a target sentence containing a masked position. The context sentence is either negative or affirmative. In the negative case, it renders semantically impossible a certain token at the masked position (\textit{sail} in example \Next), which is itself plausible in the positive case (see section \ref{ss:analyse-gubelmann}).}
\citet{gubelmann-handschuh-2022-context} have thus proposed a test where the inputs supplied to the models are self-contained (in our terminology): a context sentence is followed by a target sentence containing a masked position. The context sentence is either negative or affirmative. In the negative case, it renders semantically impossible a certain token at the masked position (\textit{sail} in example \Next), which is itself plausible in the positive case (see section \ref{ss:analyse-gubelmann}). 

\ex. Jessica is an architect who doesn’t like to sail. However, she does like to [MASK]. \label{ex-0}

\citet{gubelmann-handschuh-2022-context} observe a variable sensitivity to negation depending on the models tested, suggesting that the truth-value inversion effect of negation is more or less captured depending on the model.

In this article, we build on \citep{gubelmann-handschuh-2022-context} (hereafter \textbf{GH22}), taking up the idea of self-contained inputs, allowing us to target understanding of the semantics of negation independently of world knowledge. Our contributions are the following: 
\begin{itemize}
    \item a finer-grained analysis of GH22 experiments, uncovering a much more contrasted picture. In particular, averaged results for different input patterns mask significant sensitivity to factors other than negation (e.g., an intensifier \textit{really} or \textit{does}).

    \item the development of a more controlled test\footnote{\it https://github.com/davkletz/self-neg-test}, using self-contained inputs organized in minimal pairs differing only in polarity, as well as the introduction of control tests (double negation, use of a non-negative adverb instead of \textit{not}, and variations on coreference between NPs in the context sentence and the target sentence). 
\end{itemize}

Finally, this test enables us to make a detailed assessment of four models, and to conclude that among these, only \texttt{roberta-large} reasonably meets the defined criteria. Crucially, a number of models like \texttt{bert-large} seemed reasonably sensitive to negation in GH22, do pass our baseline test, but don't pass the control tests at all, calling into question the positive interpretation of the baseline test. This highlights the many limitations to PLMs' understanding of negation for English, and the need for highly controlled tests to reach solid conclusions. 

\section{Replication of \citep{gubelmann-handschuh-2022-context}}
\label{s:replication}

\subsection{Presentation of the test}

The GH22 test aims at studying the tokens predicted at a masked position, within an input consisting of two sentences, a C(ontext) sentence followed by a T(arget) sentence. The actual examples provided as input to PLMs are obtained by instantiating variables within patterns:
the context sentence C contains a variable ACT, to be instantiated with a verb (called \textbf{ACT-token}), like \textit{sail} in \ref{ex-1}, embedded in a negative (\textit{doesn't like to ACT}) or affirmative (\textit{tries to ACT as often as possible}) phrase. 
We refer to \textbf{Cn} and \textbf{Cp} as the negative and affirmative contexts. The sentence T contains a masked position, 
%is always affirmative,  %pA: ça me semble trompeur de dire que c'est toujours affirmatif
and is defined in such a way that repetition of the ACT-token is acceptable with an affirmative context (Cp), while semantically impossible with a negative context (Cn): for example, repeating \textit{sail} in the masked position is plausible in~\ref{ex-2} while semantically impossible in~\ref{ex-1}. 

   \ex. NAME(Jessica) is PROF(an architect) who \textbf{doesn’t like} to ACT(sail). \uline{However}, PRON(she) \uline{does} like to [MASK]. \label{ex-1}

   \ex. NAME(Jessica) is PROF(an architect) who \textbf{tries} to ACT(sail) \textbf{as often as possible}. \uline{So}, PRON(she) \uline{really} \textbf{likes to} [MASK]. \label{ex-2}

The metric proposed by GH22 is the rate of repetition of the ACT-token (\textbf{\%-ACT-repetition}), \textit{i.e.}\ the percentage of instantiated examples for which the top-1 at position MASK is the ACT-token itself.
In the Cp case, a high repetition rate is acceptable, as the ACT-token is not mandatory at this position, but plausible. In the Cn case, a high \%-ACT-repetition is clearly a sign of a failure of the model, by construction of the input examples. Note, however, that a weak \%-ACT-repetition may be due to a good ``understanding'' of negation by the model, but may also stem from inconsistencies, if the model predicts agrammatical tokens in top-1 in this context.

GH22 also varied other parameters, such as the presence or absence of ACT-coordinated verbs in C\footnote{As in \textit{NAME is a PROF who doesn't like to ACT, ACT1 or ACT2.}}, intensifiers \textit{does} and/or \textit{really} in T, and a discourse connective in T (contrastive \textit{however} if C is negative, mplicative \textit{so} if C is affirmative). Details of the combinations tested by GH22 are given Table~\ref{tab:gubelman_combinations_summmary}.

\begin{table}[!htb]
\centering
\begin{tabular}{|c|cccc|}
\hline
pol. & main v in C & aux & adv & conn.\\
\hline
N & doesn't like to & \checkmark & - & \checkmark\\
N & doesn't like to & \checkmark & - & - \\
N & doesn't like to & - & - & - \\
P & tries to & \checkmark & \checkmark & \checkmark\\
P & tries to & - & \checkmark & - \\
\hline
\end{tabular}

\caption{Details of the parameter combinations tested by GH22. Columns: \textit{pol.} is the polarity of the context (negative or positive);  \textit{main v in C} indicates which verb was used in the C sentence; \textit{aux} (resp. \textit{adv}) indicates whether \textit{does} (resp. \textit{really}) is used in T; \textit{conn.} indicates whether T begins with a connective (contrastive \textit{however} for Cn, and implicative \textit{so} for Cp). In addition, in GH22, all these configurations are combined with the gender of the subject proper noun (fem/masc) and with 0, 1 or 2 verbs coordinated to ACT.}
\label{tab:gubelman_combinations_summmary}
\end{table}

\subsection{Pattern instantiation}
\label{ss:analyse-gubelmann}

The authors generated the input examples by instantiating first the variables NAME (with typically feminine or masculine first names), PROF (with a profession), and PRON (third person pronoun, same gender as NAME). Then, for any instantiated triplet (NAME, PROF, PRON), the ACT-token is chosen by considering the tokens predicted at the masked position in the sentence \ref{ex:external-pattern-GH22}:
%\textit{NAME is PROF and PRON likes to [MASK]}: 
either the first one, the 50th, 100th or 200th in the predicted distribution. As a consequence, the examples that will be given as input vary from one model to another, the ACT-token being adapted for each triplet and model).

\ex. \label{ex:external-pattern-GH22}
\textit{NAME is PROF and PRON likes to [MASK]}

\subsection{Critical analysis}

The reason we give these details is that a careful examination of the data set and a replication of the GH22 experiments reveal variance in the results and certain asymmetries, making it difficult to draw firm conclusions.

Firstly, the GH22 test is not organized on the basis of minimal pairs varying only in polarity (like the pair \ref{ex:Robin} above, from \citet{kassner_negated_2020}). Such minimal pairs cannot be formed, firstly because the parameter combinations are not exactly the same for cases with positive context (Cp) and those with negated context (Cn), and secondly because the embedding verb is different in Cp and Cn (\textit{tries to ACT as often as possible} versus \textit{doesn't like to} ACT). So it's hard to tell whether the variations in \%-ACT-repetition are due to negation sensitivity or to other parameters. 

These parameters (connectives and intensifiers) have a major impact on the interpretation of the examples, and more specifically on the discourse link between the context and target sentences.
The test is based on: 
\begin{itemize}
    \item examples with an affirmative context, for which a repetition of the ACT is expected and corresponds to a discourse relation `elaboration' (for instance, in the context ``\textit{Jessica is an architect who tries to dance as often as possible}'', the second sentence ``\textit{She likes to dance}'' goes in the same direction);
    \item examples with a negative context, for which a \textbf{non}-repetition of ACT is expected, corresponding to a `contrast' between C and T.
\end{itemize} 

In GH22, the interpretation of the discourse relation between C and T is supported by various clues in addition to the absence or presence of negation in C: (i) the possible co-reference between NAME and PRON, 
(ii) the semantic link between the main predicates in C and in T (e.g. for Cp cases, the link between \textit{try to ACT as often as possible} and \textit{like to ACT})
and (iii) the intensifiers \textit{does} and \textit{really} and the discourse connectives. Because they make the elaboration or contrast relation explicit, connectives make the test easier, and weaken the possibility of analyzing the models' ``understanding''  of negation. Intensifiers strengthen the elaboration relation in the positive case, but the effect is more ambiguous in the negative case.
%(iv) the absence or presence of negation on these predicates.  
This excessive number of parameters weakens the interpretation that can be made of this test.

And indeed, while overall the PLMs tested show a sensitivity to negation in the GH22 results (\textit{i.e.}\ the repetition rate is lower for Cn cases than for Cp), in replicating their experiments we observed significant variance depending on the various parameters cited above, notably the presence of coordinated verbs in C, and the rank for the choice of ACT-token (GH22 give results aggregating ranks 1 and 50). In the next sub-section, we present our replication of GH22 in detail, before moving on in the next section to our proposal for a more controlled test, based on minimal pairs varying only in polarity, and on additional control tests to ensure a correct interpretation of the results.

\subsection{Partial replication}\label{ss:resultats_replications}

In this sub-section, we present our replication results for GH22. To focus on the ways models deal with negation, we have ignored a number of parameters, and systematized the combination of retained parameters. More specifically, we have limited ourselves to (i) patterns with no coordination in context sentences and (ii) instantiations using rank-1 ACT-token (we observed significant variations with respect to these parameters). We also discard patterns with connectives, which allows us to reconcile the affirmative and negative versions of the tested inputs (since the connectives differ along with the polarity of C), and above all to remove cues favoring or hindering the repetition of the ACT.

To sum up, the parameters we have kept for this replication are the presence/absence of negation in C, the presence/absence of the intensifiers \textit{does} and \textit{really} in T, and we test all 8 combinations. 

We apply this reduced test to the models \texttt{bert-large-cased} \citep{devlin-etal-2019-bert} and \texttt{roberta-large} \citep{roberta-liu}. The results are summarized in Table~\ref{tab:replic_gubelman}.  

\begin{table}[htb]
\centering
\begin{tabular}{|r|c|cc|cc|}
\hline
n° & pol. & aux & adv & \texttt{roberta-l} & \texttt{bert-l} \\
\hline
  \multirow{2}{*}{1} & P & \multirow{2}{*}{-} & \multirow{2}{*}{-} & 44.2 & 24.6 \\
  & N &  &  & 27.3 & 3.3\\

\hline
 \multirow{2}{*}{2}  &  P & \multirow{2}{*}{\checkmark} & \multirow{2}{*}{-}   & 31.8 & 91.8\\
 & N  & & & 25.1 & 58.3\\

\hline
\multirow{2}{*}{3}   &  P & \multirow{2}{*}{-} & \multirow{2}{*}{ \checkmark}   & 94.1 & 99.6\\

&  N & &   & 25.3 & 73.6\\

\hline
\multirow{2}{*}{4} & P & \multirow{2}{*}{\checkmark} & \multirow{2}{*}{\checkmark} & 99.8 & 100\\
 & N &&  & 55.9 & 96.3\\

\hline
\end{tabular}

\caption{\%-ACT-repetition rates, for the two models \texttt{roberta-large} \& \texttt{bert-large}, using GH22 patterns without coordination in the context sentence C nor discourse connectives. As in GH22, the ACT-token is chosen as the top-1 prediction for \textit{NAME is a PROF and PRON likes to [MASK]}. Columns: \textit{pol}: polarity in C; \textit{aux}: presence of \textit{does} in T; \textit{adv}: presence of \textit{really} in T.}
\label{tab:replic_gubelman}
\end{table}

The results are analyzed by comparing the P lines with their corresponding N lines, and considering the drop in repetition %rate when moving to N
(drop = P rate $-$ N rate). GH22 consider that the greater the drop, the more sensitive the model is to negation.
For both models, a drop is indeed observed for all 4 pairs of P/N patterns, so we can say that the test is effective. Note, however, that for \texttt{bert-large-cased}, the drop is small in patterns with \textit{really}. The model seems to interpret intensifiers as elaborations, and doesn't seem to be able to interpret a contrast despite the negation in the context sentence (\textit{NAME is a PROF who doesn't like to ACT. PRON really does like to [MASK]}). 

These results underline the strong interference of intensifiers on ACT repetition rates and drops, yet only a pattern without intensifiers (or connectives) comes close to a minimal pair targeting negation.

In addition, we believe that the method has a significant shortcoming. For patterns with positive polarity, the ACT repetition rate can be far from 100\% (\textit{e.g.}, $31.8$ for \texttt{roberta-large}, pattern 2P). For this pattern, therefore, $100-31.8=68.2$\% of the affirmative examples are such that the top-1 prediction is not the ACT. This corresponds to cases like \ref{ex:defaut-P}). 

\ex.  Maria is a doctor who tries to \textbf{pad} as often as possible.\\She does like to [{\small MASK}]top-1=\textbf{teach}.\label{ex:defaut-P}

However, in this case, GH22 count a non-repetition in the corresponding negative example (as in example \ref{ex:defaut-No}) as a proper handling of negation by the model. 
\ex.  Maria is a doctor who doesn't like to \textbf{pad}. \\She does like to [{\small MASK}]top-1=\textbf{follow}.\label{ex:defaut-No}
%\b. She does like to [{\small MASK}]top-1=\textbf{teach}.\label{ex:defaut-Ns}

But such a non-repetition can no longer be taken as an evidence for an understanding of negation: it is not so striking not to repeat the single forbidden token (\textit{pad}), since it was not repeated in the affirmative case (as in \ref{ex:defaut-P}).
%This is inappropriate: a non-repetition in the negative case of an ACT-token which is not repeated in the affirmative case says nothing about the model's understanding of negation: either the top-1 is neither the ACT nor the top-1 of the positive example (as in \ref{ex:defaut-No}), and there is certainly an effect of negation but not necessarily an understanding of its semantics, or the top-1 is not the ACT, but is the same top-1 as in the positive case (\textit{teach}, as in \ref{ex:defaut-Ns}), and then this non-repetition is compatible with an understanding or incomprehension of the negation.

%\ex.  Maria is a doctor who doesn't like to \textbf{pad}. 

%pA: je pense qu'on n'a pas besoin du résumé 
%To summarize \textbf{instantiated examples that don't lead to a repetition in the positive case are interpreted as good negation management, when in fact they say nothing about negation management}.
In order to circumvent the shortcoming, we propose to consider by construction only examples leading to a repetition of the ACT-token in the affirmative case.

\section{Our proposal: the \ST}

\subsection{Patterns}

On the basis of the above findings, we propose to create a test that is strongly inspired by GH22, but that allows us to draw a more reliable conclusion: we want to be able to attribute any drop in the ACT-token repetition rate solely to negation, and thus judge whether a model has mastered the semantics of verbal negation.

We keep the principle of ``self-contained'' inputs, composed of a context sentence (C), and a target sentence (T) ending with a masked position, syntactically calling an infinitive verb. But we propose a single pattern for C and T, each sentence being either affirmed or negated, so that variation in C and T is limited to the presence or absence of negation. We give the two variants Cp and Cn and the two variants Tp and Tn in table \ref{tab:sctest_base_patterns}. By combining these variants, we obtain four patterns (CpTp, CpTn, but also CnTp and CnTn). 

\begin{table}[H]
\centering
\begin{tabularx}{
\linewidth}{|*2{>{\centering\arraybackslash}X|}@{}
}
\hline
Context & Target\\
\hline
\textbf{Cp}: NAME is {a} PROF who likes to ACT. & \textbf{Tp}: PRON is happy to [MASK]. \\
\hline
\textbf{Cn}: NAME is {a} PROF who doesn't like to ACT. & \textbf{Tn}: PRON isn't happy to [MASK]. \\
\hline
\end{tabularx}
\caption{Context and target sentences, either positive and negative, used for the base \ST. 
}
\label{tab:sctest_base_patterns}
\end{table}
Note that in CpTp, and to a lesser extent CnTn, although the repetition of the ACT-token is not mandatory, it leads to a pragmatically felicitous discourse. In contrast, the repetition is semantically forbidden in CpTn and CnTp.

\subsection{Instantiation of examples} %% Construction (First column of the test) }
\label{ss:construction_test}
As in GH22, final examples are obtained by instantiating NAME, PROF and ACT (PRON is \textit{she} or \textit{he} depending on the gender of the proper noun instantiating NAME), but we modify the way ACT is instantiated, to resolve the shortcoming described in section \ref{ss:resultats_replications}: we only consider by construction positive examples (pattern CpTp) leading to a top-1 repetition. To do this, instead of using a different sentence, external to the test (GH22 used \ref{ex:external-pattern-GH22}), we take the CpTp pattern (\textit{NAME is a PROF who likes to ACT. PRON is happy to [MASK].}), and for each pair [NAME, PROF], for each model, we instantiate ACT with an English intransitive verb such that the top-1 prediction at the masked position is that very same verb. 

More precisely, the instantiation procedure is as follows: we have four lists (100 female proper nouns, 100 male proper nouns, 91 professions, and a number of monotokenized intransitive verbs, the number varying according to the tokenizer of the models). For proper nouns and professions, we re-use the GH22 lists. For verbs, we use monotokenized infinitives among English verbs that may have an intransitive usage, by cross-referencing the list on this wiktionary page \url{https://en.wiktionary.org/wiki/Category:English_intransitive_verbs} and the verbs present in Verbnet \citep{KipperSchuler2006}. We apply this procedure to two bert models (\texttt{bert-base-cased}, \texttt{bert-large-cased}) and two roberta models (\texttt{roberta-base} and \texttt{roberta-large}), and we obtain 597 and 106 verbs for the \texttt{bert} and \texttt{roberta} models respectively.

For each PLM and for each of the 2*100*91=18200 [NAME, PROF] pairs, we compute the subset of verbs in the list that lead to a top-1 repetition. The number of such [NAME, PROF, verb] triplets is shown in row number 4 of table \ref{tab:stats_CpTp}. We then randomly select at most 20 verbs for each model and each [NAME,PROF] pair (row 6 of table \ref{tab:stats_CpTp}). Note that these subsets hence depend on the tested model.

Table~\ref{tab:CpTp_pred_example} illustrates the process for the pair [\textit{Jessica, dancer}]. 

\begin{table}[H]
\centering
\begin{tabular}{|l|r|c|}
\hline
\multicolumn{3}{|l|}{\textbf{Instantiated NAME/PROF}: \textit{Jessica, dancer}}\\
\hline
\multicolumn{3}{|l|}{\textbf{Tested verb}: \textit{smoke}}\\
\hline
\multicolumn{3}{|l|}{\textbf{Tested example}: \textit{Jessica is a dancer who likes}}\\
\multicolumn{3}{|l|}{\textit{to smoke. She is happy to [MASK].}}\\
\hline
    model & top 1 pred. & retained?\\
   \hline

   \texttt{bert-base-cased} & \textit{smoke} & \checkmark \\
    \hline

    \texttt{bert-large-cased} & \textit{smoke}& \checkmark \\   
    \hline

    \texttt{roberta-base} & \textit{dance} & no\\
    \hline

    \texttt{roberta-large} & \textit{chat} & no\\
    \hline
    
    \end{tabular}
    \caption{Example of selection of [NAME, PROF, ACT] triplets, for a given instantiated [NAME=\textit{Jessica}, PROF=\textit{dancer}] pair. When instantiating ACT with \textit{smoke}, the top-1 at the MASK position is \textit{smoke} (repetition) for the models \texttt{bert-base-cased} and \texttt{bert-large-cased}, and will eventually be selected when retaining 20 random such verbs for the given input pair.
    }\label{tab:CpTp_pred_example}
    \end{table}

For each model, the triplets thus obtained to instantiate NAME, PROF, ACT are then used to form the saturated examples for each of the 
patterns.
%five patterns CpTp, CnTp, CpTn, CnTn and CpTv. 

    \begin{table*}[!hbt]
    \centering
    \begin{tabular}{l r r r r}
   
    \textbf{model} & \texttt{bert-b-c} & \texttt{bert-l-c} & \texttt{roberta-b} & \texttt{roberta-l} \\
   \toprule
   1. Available verbs &  597  &  597  & 106  & 106  
   \\
   \midrule
   2. Available NAME,PROF pairs &  18200  &  18200   & 18200   & 18200   
   \\
   \midrule
   3. Tested triplets (row 1 $\times$ row 2) (*10$^6$) & 10.9 & 10.9 & 1.9 & 1.9
   \\
   \midrule
   4. \quad  $\hookrightarrow$  leading to ACT repetition (*10$^6$) & 2.4 & 2.0 & 1.2 & 0.4
   \\
    \midrule
   5. Ratio   (row 4/row 3, \%) &  21.7 & 18.4 & 61.9 & 18.3 
   \\
      \toprule
   6. Selected triplets  &  364000 &  363922 & 362027 & 107856
   \\
   \bottomrule
    \end{tabular}
    \caption{Statistics for the selection stage of triplets instantiating NAME, PROF, ACT, for each model. \textbf{Row 3:} number of tested triplets (NAME, PROF, verb). \textbf{Row 4:} number of such triplets for which the instantiated CpTp example leads to a repetition (top-1 prediction is identical to the ACT-token). \textbf{Row 6:} number of selected triplets among those of row 4 (retaining at most twenty verbs for each [NAME,PROF] pair). 
    } \label{tab:stats_CpTp}
    \end{table*}

\subsection{Test interpretation}

For each pattern, we can detail how a decrease or stability in \%-ACT-repetition should be interpreted in relation to the rate of 100\% repetition for CpTp. As we'll always be comparing \%-ACT-repetition with the 100\% rate obtained by construction for CpTp, we prefer to consider a measure of rate decrease: \textbf{drop} = 100 $-$ \%-ACT-repetition.

\begin{itemize}
\item \textbf{CnTp}: this pattern is an evolution of patterns proposed by GH22, but designed here to form a true minimal pair with CpTp. By construction, ACT-token is semantically impossible at the masked position, so a small drop would mean that the model doesn't interpret negation in C. On the contrary, the larger the drop (the maximum being 100), the more likely it is that the model interprets correctly the negation in C. Note that any other verb is semantically and discursively possible in the masked position, and corresponds to a contrast discourse relation between Cn and Tp.  
\item \textbf{CpTn}: we also add the case where negation is in the target sentence, and therefore closer to MASK. Here again, ACT-token is semantically impossible at the masked position, and the drop interpretation is the same as for CnTp. 
\item \textbf{Control pattern CnTn}: here the repetition of the ACT-token is discursively natural. A high-performing pattern is expected to have only a marginal drop. The pattern is used to check that a negation in one sentence is correctly interpreted in relation to the polarity in the other sentence, and not just in isolation.
\item \textbf{Control pattern CpTv}: we also add a control where the modification with respect to CpTp is not the addition of the negative adverb, but the addition of another adverb, \textit{very}, in T. This pattern makes it possible to check whether a drop in CpTn is really attributable to the negation in T, and not simply to the addition of any adverb. More generally, as ACT has been instantiated to obtain 100\% repetition in the CpTp pattern, this CpTv pattern makes it possible to check the stability of ACT-token repetition: if a model's predictions are often different depending on the presence or absence of \textit{very} in \textit{NAME is a PROF who likes to ACT. PRON is (very) happy to ACT}, then this would be a sign that any change could potentially have a lot of impact, and it would prevent any positive interpretation of a drop for this model.
\end{itemize}

\subsection{Properties of the test}
These patterns have been chosen to limit the factors that can be used to interpret the discourse relation between C and T. In our case, the interpretation is solely driven by (i) the coreference between NAME and PRON, (ii) the semantic link between \textit{like to ACT} and \textit{be happy to ACT} and (iii) the absence or presence of negation on these predicates: only (iii) varies within the test, (i) and (ii) remain
stable, and no intensifier or discourse connector cues are added that would favor or hinder the repetition of ACT-token.  

In this way, we can form true minimal pairs varying only by a negation, in C or in T: for each triplet instantiating NAME, PROF and ACT, we have four minimal pairs (CpTp / CpTn), (CpTp / CnTp), (CnTn / CnTp) and (CnTn / CpTn). 

By forcing an ACT repetition rate of 100\% for the CpTp pattern, we totally avoid positive examples that don't lead to a repetition, which render the corresponding negative examples %at best 
unusable %, at worst misleading 
(cf. sub-section~\ref{ss:resultats_replications}).
What's more, the CpTp pattern now serves as a reference point%
%benchmark
, and decreases in \%-ACT-repetition are more comparable one with another, whether for a comparison between models, for the same pattern, or a for comparison between patterns for the same model.
Finally, we make sure to obtain instantiated examples where the discourse relation between \textit{like to ACT} and \textit{be happy to ACT} is ``understood''. In this way, a lower repetition rate in a negative context will be all the more significant. 

Note that the procedure to select [NAME, PROF, verb] triplets yields a large number of ACT-repetitions in CpTp (cf. the ratios for each model provided at row 5 of table \ref{tab:stats_CpTp}). This confirms that repetition in the CpTp pattern is pragmatically felicitous, although not mandatory. We observe that this ratio is much higher for the \texttt{roberta-base} model compared to the other three models. We cannot state whether this stems from a higher tendency to repeat tokens or from a preference to interpret the discourse relation between the two sentences as an elaboration. 

\subsection{Models evaluation}

We apply our test to the four above-mentioned models and provide the results in Table~\ref{tab:selfCTest}.

\begin{table*}[!h]
\centering
\begin{tabular}{|c | r r r r|}
\hline
  \textbf{Pattern} & \texttt{bert-b-c} & \texttt{bert-l-c} & \texttt{roberta-b} & \texttt{roberta-l} \\
\hline

CpTn & 3.6 & 44.7& 27.7 & 64.7   \\

CnTp & 1.2 & 16.5 & 66.9 &  82.8\\
\hline
CnTn  & 1.5 & 9.7 & 12.1 & 43.5\\

CpTv & 25.5 &  42.9 & 23.3 &  26 \\ 
\hline

\end{tabular}
 \caption{Drops of the \%-ACT-repetition with respect to 100\% for CpTp, when applying the \ST  \ to four PLMs. To pass the test, drops should be high for the first two lines, and low for the last 2.}\label{tab:selfCTest}
  \end{table*}

Recall that passing our test implies having strong drops for the CpTn and CnTp patterns, and that these drops be %are
greater than in the control pattern CpTv, and to a lesser extent in the CnTn pattern.

The \texttt{bert-base-cased} model fails the test completely: the drop is almost non-existent for the CpTn and CnTp patterns. Moreover, the drop is much stronger for the CpTv control pattern: in the context of \textit{NAME is a PROF who likes to ACT}, the model repeats ACT less often in \textit{PRON is very happy to MASK} than in \textit{PRON isn't happy to MASK}.

Although the drop of the \texttt{bert-large-cased} model is larger for the CpTn and CnTp configurations than those of \texttt{bert-base-cased}, its drop in the CpTv configuration is still too high to conclude that this model understands negation.

The \texttt{roberta-base} model shows drops closer to our expectations: its CnTn drop is smaller than those of CpTn and CnTp ($20.7$ and $46.7$). But since the drop for the CpTv control is $21.3$, only the $46.7$ drop is significant.

Finally, the model that seems to have acquired the most robust understanding of negation is \texttt{roberta-large}, having both a drop of over 50\% for CpTn and CnTp, and a small drop for the controls. The maximum drop is obtained for CnTp, i.e. with a negation in the context sentence. It remains to be investigated why this configuration is better handled than CpTn.
           
To sum up, none of the models reaches drops close to 100\% for CpTn and CnTp: many examples lead to a repetition of a token that is semantically forbidden by the context sentence. Nevertheless, it seems that the \texttt{roberta} models, and in particular \texttt{roberta-large}, ``understand'' the semantic value of verbal negation in English better than \texttt{bert}.
Moreover, within a family of models, the \texttt{large} version performs better. 

\section{Additional controls: forcing non-coreference}

In the results analyzed in the previous section, the drop can only be interpreted as an understanding of negation if the model has resolved the co-reference between the proper noun in C and the pronoun in T. In the absence of such a resolution, a repetition of the ACT is neither forbidden nor required.

While the ability of \texttt{bert} to resolve coreference has been evidenced by \citet{clark-etal-2019-bert}, we need to ensure that this resolution is effective in the case of our patterns.
To do this, instead of directly testing coreference resolution, we build a set of alternative examples to the base examples, in which non-coreference is forced.
In practice, we replace the pronoun in T by a proper noun other than the one used in C, with two variants, depending on whether or not these two proper nouns have the same gender (cf. examples 1 and 2 table~\ref{tab:ex_non-coref}).  
If the model does indeed resolve coreference in base examples, then we should observe a much smaller %larger ERREUR TRAD!
drop for examples with forced non-coreference: in the absence of coreference, the context sentence no longer gives information about the target sentence, and therefore no longer prohibits or favors the choice of a particular token. As the sequences have been selected to favor repetition of the ACT token, this repetition should however remain high.

We also consider cases where we help the model establish a co-reference, so as to test only the impact of negation, independently of the models' ability to establish the co-reference between the proper noun and the pronoun in the basic examples. To this end, we use the same proper noun in C and T (cf. example 3 table~\ref{tab:ex_non-coref}). The repetition gives a less natural example, but in which the coreference is forced. 
% décision : non \TBD{do we add this argument or leave it out? This also allows us to check that the non-coreference test is not polluted by the use of 2 first names instead of one first name and then a pronoun.}\parm{à mon avis on laisse tomber, ce n'est pas crucial et il faut prendre plus de temps pour expliquer.}

\begin{table*}[!h]
\centering
\begin{tabular}{lccc}

id & Type & Context & Target \\
\toprule
1 & non coref, same gender & Joyce is a designer who likes to smoke.& Janet really likes to [MASK].
\\
2 & non coref, other gender & John is a dentist who likes to dance.&  Anna really likes to [MASK].
\\
3 & forced coref by repetition & Judith is a diplomat who likes to drink.& Judith really likes to [MASK].
\\
\bottomrule
\end{tabular}
\caption{Examples of the corefence control test. In 3, coreference is forced by using the same name in C and T. In 1 and 2, coreference is ruled out by using distinct names, of either same or different genders.}\label{tab:ex_non-coref}
\end{table*}
    
Triplets are selected using the same procedure as in section \ref{ss:construction_test}, namely retaining only triplets leading to a top-1 repetition for the CpTp pattern, and at most 20 verbs for a given [NAME,PROF] pair.

Results for \texttt{roberta} models are provided in table~\ref{tab:coref_ST_roberta} (those for \texttt{bert} models are in appendix~\ref{app:non-coref}, table \ref{tab:coref_ST_bert}). A first observation is that the number of selected triplets (first row of tables~\ref{tab:coref_ST_roberta} and \ref{tab:coref_ST_bert}) undergoes a severe decrease.
This is consistent with the fact that in such configurations, repetitions are pragmatically less felicitous.

An efficient model is expected to obtain small drops for the Non-Coref columns, while retaining large drops in the Coref column, for the CpTn and CnTp cases. The drops in CnTn and CpTv controls should remain small.

\begin{table}[H]
    \begin{subtable}[h]{0.45\textwidth}
        \centering
        \begin{tabular}{|c | c  |c c|}
            \hline
               \textbf{Pattern} & Coref  & \multicolumn{2}{c|}{Non-Coref}\\
                & & 
               {Same-gend.} & {Other-gend.} \\
            \hline

\# ($\times 10^3$) &118.7 & 39.6 & 44.7\\
            
            \hline 
    CpTn &  3.9 &  7.5 & 6.9 \\ 
    CnTp & 12.6 &  3.3 & 3.4 \\ \hline
    CnTn & 1.5 & 3.4 & 2.4 \\ 
    CpTv & 8.7 &  4.5 & 4.8 \\

            \hline

        \end{tabular}
       \caption{\texttt{roberta-base}}
       \label{tab:week1}
    \end{subtable}
    \hfill
    \begin{subtable}[h]{0.45\textwidth}
        \centering
       \begin{tabular}{|c |  c | c c |}
\hline
     \textbf{Pattern} & Coref  & \multicolumn{2}{c|}{Non-Coref}\\
                & & 
               {Same-gend.} & {Other-gend.} \\

\hline

\# ($\times 10^3$) &60.8 &4.4 &5.1 \\

\hline

CpTn &  28.9 &  11.7 & 12.1 \\ 
CnTp &  64.1 &  1.9 & 10.8 \\ \hline
CnTn & 14.3 &  6.2 & 7.1 \\

CpTv & 17.3 &  9.4 & 11.5 \\
 
\hline

\end{tabular}
        \caption{\texttt{roberta-large}}
        \label{tab:non-coref-roberta}
     \end{subtable}
     \caption{{\bf Last 4 rows}: drops of the \%-ACT-repetition for the \texttt{roberta} models, when forcing coreference by using the same name in C and T (\textbf{Coref}) or forcing non-coreference using different names (\textbf{Non-Coref}), either of same or different genders. {\bf First row (\#)}: number of selected [NAME, PROF, ACT] triplets, among those leading to a top-1 repetition in the CpTp pattern (still retaining at most twenty verbs for each [NAME,PROF] pair).}\label{tab:coref_ST_roberta}
\end{table}

We can see this is not the case for the \texttt{roberta-base} model: the drops in the upper left part of table \ref{tab:coref_ST_roberta}a are smaller with respect to the ``vanilla'' examples (with a pronoun in T sentences). It is as if the model interpreted the same two names as non-coreferent. It is though impossible to conclude whether this is the case (in which case the smaller drops do not mean that negation is misunderstood), or whether the model interpreted the coreference correctly, but failed to interpret negation.

On the other hand, the trends observed for \texttt{roberta-large} (table \ref{tab:coref_ST_roberta}b) do follow our expectations: the drops do remain large for the Coref case for CpTn and CnTp (and significantly larger than for the CnTn and CpTv controls) but they are small for the non-coreference patterns. This confirms the observations made for this model with the previous test, and thus further confirms the abilitiy of this model to capture the semantics of verbal negation.

\section{Conclusion}

In this paper we propose a methodology and a dataset to study PLMs' abilities to correctly interpret the semantics of negation, more precisely verbal negation in English. We were inspired by \citet{gubelmann-handschuh-2022-context}, who proposed \textit{self-contained} examples, consisting of two sentences, the first serving as a context that favors or hinders the repetition of a certain verb in the second sentence.
After critically analyzing this test, we propose an improved version, which is more controlled, more systematic, and entirely based on examples forming minimal pairs varying only in the presence or absence of verbal negation. We have sought to minimize the interpretations that the models have to make in addition to the negation interpretation, so that the observed results can be more reliably interpreted as the model's good or bad ``understanding'' of negation. 

We applied our test to four pretrained Transformer-based language models. A detailed analysis of the results shows a continuum of situations: \texttt{bert-base} is globally unable to take verbal negation into account, \texttt{bert-large} is a little better at first sight, but the control tests we added show its limitations. \texttt{roberta-base} partially passes the basic test, but is disappointing when it comes to controlling co-reference resolution. Only the \texttt{roberta-large} model shows trends in line with expectations, for both base and control patterns, clearly showing some ability to capture the semantics of verbal negation in English.

However, for all the models we tested, a significant number of examples get a top-1 prediction that is exactly the token semantically forbidden by the context. This shows how much room for improvement remains for this type of models. % ERREUR de TRAD! "modèle" a été traduit par "pattern"

We chose to focus on verbal negation, being the most frequent form of negation in English, but we plan to extend our test to other forms of negation. Extension to other languages is also considered.

\section{Limitations}
The \ST{}  only works on a Masked language modeling task. As such it is clearly designed for bidirectional models. 
Applying it to generative language models would require a complete rethinking of the test.

\bibliography{anthology,custom}
\bibliographystyle{acl_natbib}

\newpage

\appendix

\section{Results non coreference for \texttt{bert} models}
\label{app:non-coref}

\begin{table}[H]
    \begin{subtable}[h]{0.45\textwidth}
        \centering
        \begin{tabular}{|c |c  |c c|}
            \hline
              \textbf{Pattern} & Coref  & \multicolumn{2}{c|}{Non-Coref}\\
                & & 
               {Same-gend.} & {Other-gend.} \\

            \hline
                \# ($\times 10^3$) &21.4 &7.2 & \\

            \hline

           CpTn &  2.0  & 3.4 & 2.9 \\
           CnTp &  2.2 & 2.3 & 2.3 \\
           CnTn &  1.4 & 4.1 & 3.7 \\
           CpTv &  10.3 &  13.1 & 15.3 \\
            \hline

        \end{tabular}
       \caption{\texttt{bert-base}}
       \label{tab:week1-bert}
    \end{subtable}
    \hfill
    \begin{subtable}[h]{0.45\textwidth}
        \centering
       \begin{tabular}{|c | c | c c|}
\hline
   
     \textbf{Pattern} & Coref  & \multicolumn{2}{c|}{Non-Coref}\\
                & & 
               {Same-gend.} & {Other-gend.} \\

    \hline
    
    \# ($\times 10^3$) &47.1 &14.3 & 16.5 \\
    
    \hline

CpTn &  41.1  & 47.9 & 44.7 \\ 
CnTp &  5.8  & 6.4 & 5.7 \\ 
CnTn &  2.9  & 7.3 & 5.8 \\
CpTv & 38.5  & 44.8 & 45.7 \\

\hline

\end{tabular}
        \caption{\texttt{bert-large}}
        \label{tab:non-coref-bert}
     \end{subtable}
     \caption%{Drops of the \%-ACT-repetition for the \texttt{bert} models, when forcing coreference by using the same name in C and T (\textbf{Coref}) or forcing non-coreference using different names (\textbf{Non-Coref}), either of same or different genders. \DK{The "\#" line indicates the number of triplets that have been selected for repetition in CpTp (still retaining at most twenty verbs for each NAME, PROF pair)}}
     {{\bf Last 4 rows}: drops of the \%-ACT-repetition for the \texttt{bert} models, when forcing coreference by using the same name in C and T (\textbf{Coref}) or forcing non-coreference using different names (\textbf{Non-Coref}), either of same or different genders. {\bf First row (\#)}: number of selected [NAME, PROF, ACT] triplets, among those leading to a top-1 repetition in the CpTp pattern (still retaining at most twenty verbs for each [NAME,PROF] pair).}\label{tab:coref_ST_bert}
\end{table}

\end{document}